\documentclass[letterpaper]{article} 
\usepackage{aaai23}  
\usepackage{times}  
\usepackage{helvet}  
\usepackage{courier}  
\usepackage[hyphens]{url}  
\usepackage{graphicx} 
\usepackage{natbib}  
\usepackage{caption} 
\usepackage{algorithm}
\usepackage{algorithmic}

\usepackage{times}
\usepackage{soul}
\usepackage{url}
\usepackage{multirow}
\usepackage{amsthm}
\usepackage{subcaption, booktabs}

\usepackage{newfloat}
\usepackage{listings}
\DeclareCaptionStyle{ruled}{labelfont=normalfont,labelsep=colon,strut=off} 
\floatstyle{ruled}
\newfloat{listing}{tb}{lst}{}
\floatname{listing}{Listing}
\title{Multimodal and Explainable Internet Meme Classification}
\author{
Abhinav Kumar Thakur$^{1}$,
Filip Ilievski$^{1}$,
Hông-Ân Sandlin$^{2}$,
Zhivar Sourati$^{1}$,
Luca Luceri$^{1}$,
Riccardo Tommasini$^{3}$ and
Alain Mermoud$^{2}$
}
\affiliations{
$^1$University of Southern California, Information Sciences Institute, USA\\
$^2$Cyber-Defence Campus, armasuisse Science and Technology, Switzerland\\
$^3$Institut National des Sciences Appliquées, France\\
\{akthakur, ilievski,souratih, lluceri\}@isi.edu,
\{hongan.sandlin, alain.mermoud\}@ar.admin.ch,
riccardo.tommasini@insa-lyon.fr
}

\begin{document}

\maketitle

\begin{abstract}

In the current context where online platforms have been
effectively weaponized in a variety of geo-political events and
social issues, Internet memes make fair content moderation at scale even
more difficult. 
Existing work on meme classification and tracking has focused on black-box methods that do not explicitly consider the semantics of the memes or the context of their creation.
In this paper, we pursue a modular and explainable architecture for Internet meme understanding.
We design and implement multimodal classification methods that perform example- and prototype-based reasoning over training cases, while leveraging both textual and visual SOTA models to represent the individual cases. We study the relevance of our modular and explainable models in detecting harmful memes on two existing tasks: Hate Speech Detection and Misogyny Classification. We compare the performance between example- and prototype-based methods, and between text, vision, and multimodal models, across different categories of harmfulness (e.g., stereotype and objectification). We devise a user-friendly interface that facilitates the comparative analysis of examples retrieved by all of our models for any given meme, informing the community about the strengths and limitations of these explainable methods.
\end{abstract}

\section{Introduction} 

The moderation of content on social media is becoming one of the main societal challenges as online platforms have been effectively weaponized in a variety of geo-political events and social issues \cite{pierri2022does,nogara2022disinformation,chen2022charting,pierri2022online}. While lowering barriers to information sharing can guarantee freedom of expression, research showed that it also facilitates the diffusion of harmful narratives, including violent content and misinformation \cite{badawy2018analyzing,bessi2016social,tahmasbi2021go}. The detection of harmful content is challenging, given that content
can be easily created in different modalities, ranging from text to
multimedia content, and spread very quickly, sometimes amplified by coordinated accounts involved
in influence operations (e.g., bots and trolls) \cite{luceri2020detecting,zannettou2019let,luceri2019red,shao2018spread}, and often across platforms with different degrees and
strategies for moderation~\cite{Starbird2019,zannettou2019disinformation}.
Meanwhile, determining toxicity, or inappropriateness broadly is non-obvious even for humans, as social media interactions are integrated into both the virtual and the real-world context.



Content moderation policies, or the lack thereof, can have serious implications on individuals, groups, and society as a whole.
On the one hand, content moderators may react late, inconsistently, or unfairly, thus angering users~\cite{Habib_Nithyanand_2022}, as well as contributing to reinforcing and exacerbating conspiratorial narratives \cite{chen2022charting,luceri2021social}.
On the other hand, minimal content moderation may permit coordinated influence operations~\cite{diresta2019potemkin} or enable the spontaneous formation of toxic and dangerous communities, e.g., the study by \citeauthor{mamie2021anti} demonstrates how ``the Manosphere'', a conglomerate of men-centered online communities, may serve as a gateway to far-right movements. 
A recent study~\cite{delisle2019large} revealed worrying patterns of online abuse, estimating 1.1 million toxic tweets sent to women over one year. Their study also reveals that black women were 84\% more likely 
than white women to experience abuse on the platform. These studies collectively show that sexism and misogyny are still prevalent all over the globe~\cite{khan_2021}, despite initiatives such as the UN Sustainable Development Goals~\cite{biermann2017global} that emphasize the importance of gender equality, peace, and justice. 
The recent explosion of multimedia content, in the form of \textbf{Internet memes (IMs)}, makes content moderation even more difficult, especially when the context is not taken into account. 
An Internet meme can be roughly defined as ``a piece of culture, typically a joke, which gains influence through online transmission''~\cite{davison20129}. An Internet meme is based on a medium, typically an image representing a well-understood reference to a prototypical situation within a certain community.
Given that IMs are potential vectors for misinformation, political propaganda, and hate speech, enabling their scalable analysis is essential. Nonetheless, the automated analysis of IMs is challenging because of their nature: IMs are multimodal, i.e., they combine visual and language information creatively. Notably, IMs are not just funny; they are relatable and, thus, they are community-dependent. Therefore, their correct interpretation passes from the identification of the right virtual context. Moreover, IMs are succinct, i.e., they spread complex messages with a minimal information unit that connects the virtual  circumstances to the real ones. Finally, IMs are fluid, i.e., they are subject to variations and alterations. In one study by Meta~\cite{adamic2014evolution}, 121,605 different variants of one particular meme were posted across 1.14 million status updates.  
The inaccurate classification of
memes can lead to inadequate moderation interventions (removal, flagging, demotion, etc.) that,
combined with the lack of tracing mechanisms across platforms, has the potential to further decrease
public trust in social media platforms, and related moderation policies. 

Existing work on meme tracking and classification has focused on their temporal spread over time (i.e., virality)~\cite{marino2015semiotics,taecharungroj2014effect,ling2021dissecting} and high-level categorization tasks like hate speech detection that focus on perceptual features~\cite{kiela2021hateful,fersini2022semeval}.
Little work has focused on the aspects of semantics and pragmatics of a meme, which require precise feature extraction from images and from the text. Moreover, memes assume rich background knowledge about the spatio-temporal and cultural context in which they came into existence. Combining text, vision, and extra knowledge is an AI-complete problem.

In this paper, we explore explainable multimodal methods for IMs classification. 
We rely on the general idea of case-based reasoning, where a method prediction can be traced back to similar memes that the method has observed at training time. Considering the complex nature of Internet memes, we opt for case-based reasoning because it can provide transparent insights into the model reasoning, while still leveraging the representation learning ability of state-of-the-art (SOTA) models. Our contributions are:

\begin{enumerate}
    \item We build on prior work to employ explainable reasoning methods for meme understanding. These methods perform example- and prototype-based reasoning over training cases, while leveraging both textual and visual SOTA models to extract features for the individual cases.
    \item We study the relevance of our modular and explainable models in detecting harmful memes on two tasks: Hate Speech Detection and Misogyny Classification. We compare the performance between example- and prototype-based methods and between text, vision, and multimodal models, across different categories of harmfulness (e.g., stereotype and objectification).
    \item We devise a user-friendly interface that facilitates the comparative analysis of examples retrieved by all of our models for any given meme. We leverage the user interface to understand the ability of different explainable models to retrieve useful instances for case-based reasoning and inform future work about these methods' strengths and limitations.
\end{enumerate}

We make our code available to facilitate future research on explainable IM classification for social good.\footnote{\url{https://github.com/usc-isi-i2/meme-understanding}}

\begin{figure}[!t]
        \centering
        \includegraphics[width=\linewidth]{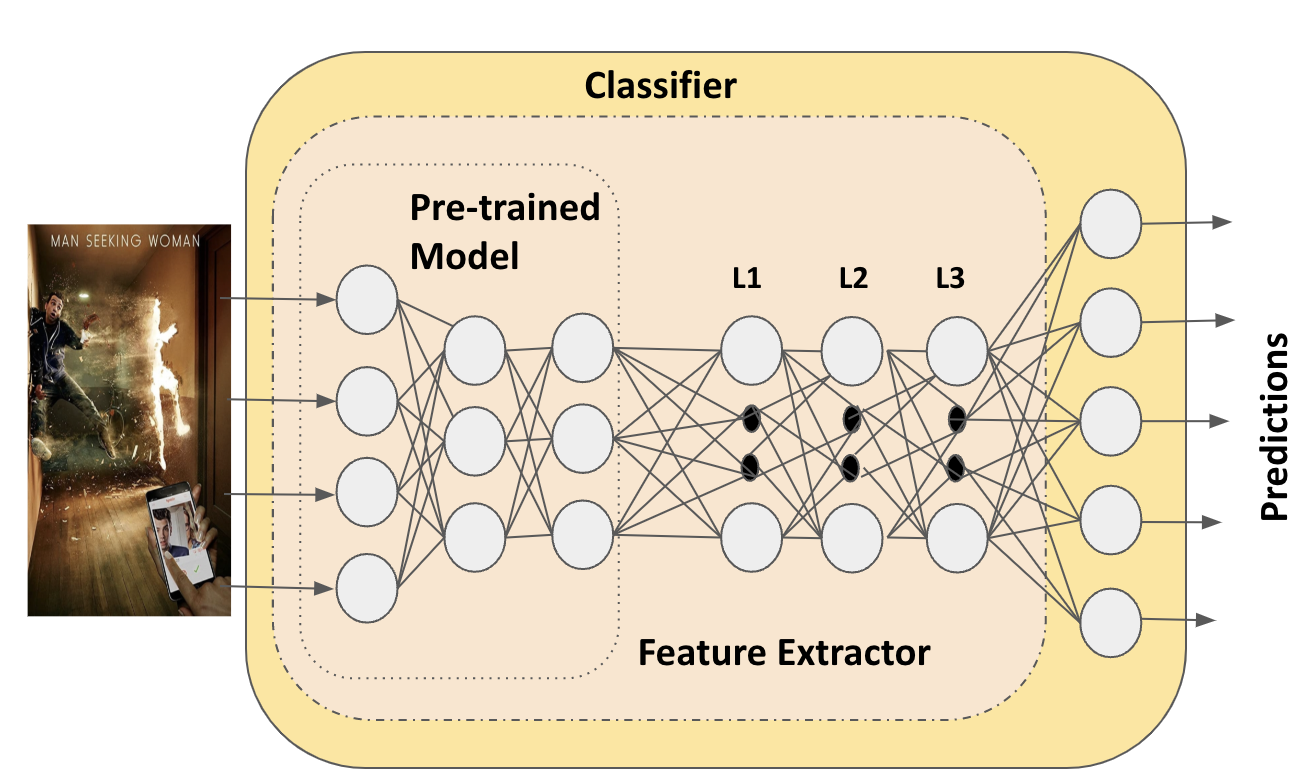}
        \caption{Classification and feature extraction model within the Example-based explanation model.}
        \label{fig:classification-model}
\end{figure}

\section{Method} 

A meme classification model that detects offensive or inappropriate memes can be easily trained. However, the black-box nature of ML models makes it difficult to interpret why a meme is flagged \cite{ANDREWS1995373}, especially when flagged wrongly. We adopt example-based and prototype-based approaches to make explainable predictions for internet meme classification tasks. Both approaches utilize a frozen pre-trained model to extract features from a meme in a transfer learning setup with a separate downstream classification model, which leverages the features to make a final decision. The modularity of
the approach enables an easy comparison over the combination of the pre-trained model and the explanation method used. We further develop a web-based visualization tool to study these explanation methodologies.

\begin{figure*}[!t]
        \centering
        \includegraphics[width=0.7\textwidth]{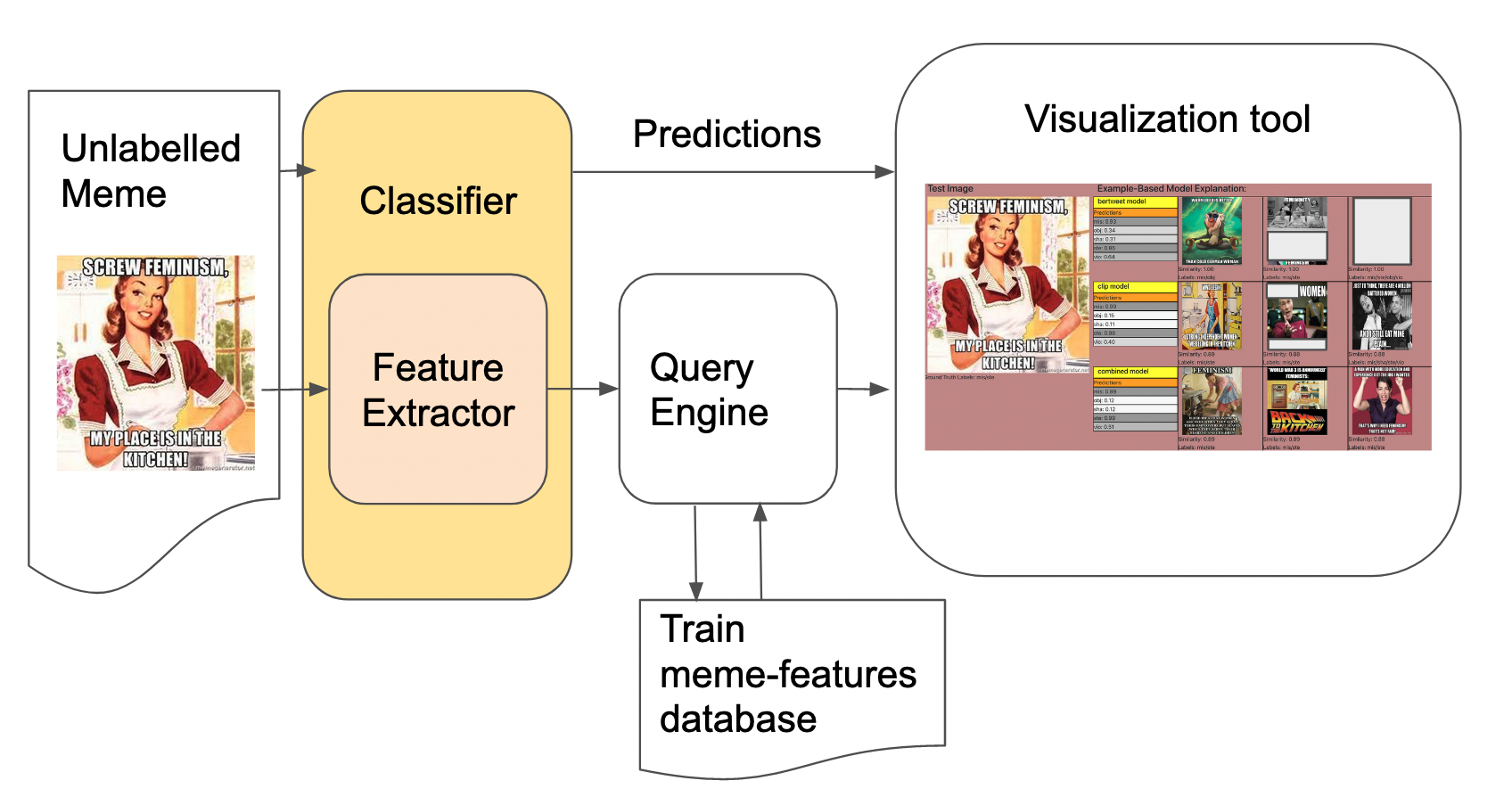}
        \caption{Example-based explanation based on similarity-based meme search. The Train meme-features database contains pre-computed features using the Feature Extractor module.}
        \label{fig:pipeline}
\end{figure*}


\subsection{Example-based Meme Classification}

We adopt an example-based method~\cite{https://doi.org/10.1111/cogs.12086} to make predictions and explain them by displaying similar memes to end-users. 
Example-based explanation works by showing training examples that have a similar representation to the test example from the model's point of view to act as a proxy to understand the model's behavior. We use example-based classification because it helps users to understand how the classification model represents a meme compared to the training dataset supporting the model prediction. 
Although Internet memes involve text, image, and often need commonsense or cultural-specific knowledge to be well understood, and that might limit the efficacy of example-based explanation, still even as a heuristic, similar examples can help the end-users understand the reasoning that is done by the model \cite{renkl2009example}.
This approach further helps analyze misclassifications and detect latent biases in the dataset \cite{benefits-example-based-explanation}.

Figure~\ref{fig:classification-model} shows the meme classification model, which applies a classification head (L1, L2, L3 and Predictions layer) over the frozen pre-trained model for prediction (see the last Subsection about Pretrained models). 
The last hidden state (output of L3) of the trained classifier is used as the extracted features for calculating the similarity between memes using cosine similarity.  Then, for an unlabelled meme, we predict the labels using this classification model. The features can be fed into a query engine to select similar images (Figure \ref{fig:pipeline}) from a database that stores pre-computed features corresponding to the training memes. To display the retrieved similar memes in a user-friendly way, we develop a visualization tool to display the model-wise predictions and similar memes from the training dataset, thus supporting the predictions with example-based explanations. 

\begin{figure*}[!t]
\centering
\includegraphics[width=0.9\linewidth]{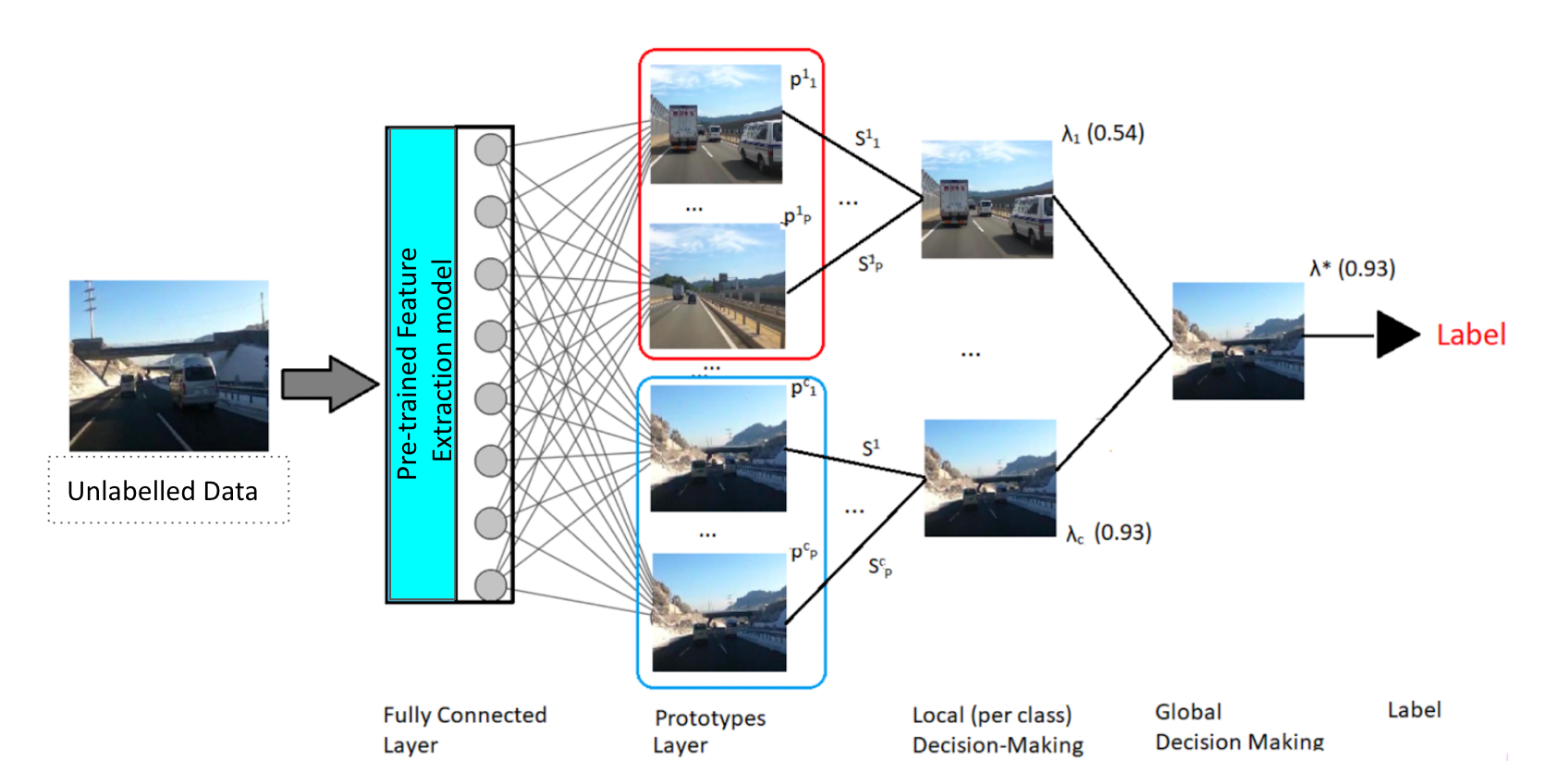}
\caption{Our architecture for prototype-based explainable classification called Explainable Deep Neural Networks (xDNN). Figure reused from~\protect\cite{https://doi.org/10.48550/arxiv.1912.02523}.}
\label{fig:xDNN}
\end{figure*}

\subsection{Prototype-based Meme Classification} 
Unlike an example-based explanation, a prototype-based explanation is not a post-hoc approach. Instead, prototype-based classification relies on learning label-wise prototypes from the training dataset followed by a rule-based decision algorithm for the classification, which makes these models inherently interpretable. 
Prototype-based explanation is based on prototype theory \cite{ROSCH1973328}, which is a theory of categorization in psychology and cognitive linguistics, in which there is a graded degree of belonging to a conceptual category, and some members are more central than others. In prototype theory, any given concept in any given language has a real-world example that best represents this concept, i.e., its \textit{prototype}. Like example-based explanation, prototype theory is also an instance of case-based reasoning, and
there has been some controversy over the superiority of one over the other. There are both claims about the superiority of prototypical examples over normal examples \cite{Johansen2005-nu}, as well as their counterparts \cite{Medin1978-je} who state that a context theory of classification, which derives concepts purely from exemplars works better than a class of theories that included prototype theory.

We reuse the implementation of Explainable Deep Neural Networks (xDNN)~\cite{https://doi.org/10.48550/arxiv.1912.02523}. It uses the training data features extracted using the pre-trained model to create class-wise prototypes (local peaks for class distribution). As shown in Figure~\ref{fig:xDNN}, the new unlabelled sample can be evaluated against these prototypes and then classified using rule-based local and global decision-making stages.



\subsection{Pretrained Models for Feature Extraction} \label{pre-tained-models}

We chose the following pre-trained models for feature extraction to analyze the information captured by models trained over different modalities and pretraining strategies.

\textbf{Textual Models:} We use the \textbf{BERT$_{base}$} model ~\cite{devlin-etal-2019-bert}, trained on BooksCorpus (800M words) and English Wikipedia (2,500M words) using two unsupervised tasks of Masked LM and Next Sentence Prediction. We expect that BERT would help analyze explainability for general-purpose formal language. Also, we used the \textbf{BERTweet} model ~\cite{nguyen-etal-2020-bertweet} having the same architecture as BERT$_{base}$ and trained using the RoBERTa ~\cite{https://doi.org/10.48550/arxiv.1907.11692} pretraining procedure over 80 GB corpus of 850M English tweets. BERTweet is supposed to be more contextually related to a meme text as tweets have short text lengths and generally contain informal grammar with irregular vocabulary, similar to IMs.

\textbf{Vision Models:} To capture visual information, we used the  \textbf{CLIP} (Contrastive Language-Image Pre-training) model ~\cite{https://doi.org/10.48550/arxiv.2103.00020}. It is trained with Natural Language Supervision over 400 million (image, text) pairs collected from the Internet with the contrastive objective of creating similar features for an image and text pair. Because of the variety of training data and unrestricted text supervision, CLIP reaches SOTA-comparable zero-shot performance over various tasks like fine-grained object classification and geo-localization, action recognition in videos, and OCR. CLIP is robust toward distribution shift between various datasets and shows better domain generalization over various datasets.

\textbf{Mixed Models:} To capture both graphical and textual information simultaneously, we concatenate features from both BERTweet and CLIP together and use them for downstream predictions.

\section{Experimental Setup} 
This section discusses the setup for evaluating our approaches over the explainable meme classification tasks.

\subsection{Tasks and Datasets}
We experimented with meme classification tasks over two existing datasets: MAMI and Hateful Memes. \\
\textbf{SemEval-2022 Task 5: Multimedia Automatic Misogyny Identification (MAMI) dataset} ~\cite{fersini2022semeval} consists of two sub-tasks of misogyny detection and its type classification. \textit{Sub-task A: Misogyny Detection Task} focuses on detecting whether a meme is misogynous. The inter-annotator Fleiss-k Agreement for sub-task A is $0.5767$.
    \textit{Sub-task B: Misogyny Type Classification Task} is a multi-class task that categorizes a meme into one or more misogyny types, namely, shaming, stereotype, objectification, and violence. A more formal description of these categories can be found in  ~\cite{fersini2022semeval}. The inter-annotator Fleiss-k Agreement for sub-task B is $0.3373$. Data statistics for both subtasks of the MAMI dataset are presented in Table~\ref{tab:dataset}. 
The inter-annotator Fleiss-k Agreement clearly shows that sub-task B is comparatively more difficult than sub-task A.

\textbf{Hateful Memes dataset} ~\cite{https://doi.org/10.48550/arxiv.2005.04790} consists of a single task of meme hate detection. The dataset consists of 10K memes equally divided into hateful and not-hateful classes; the dev and test set consist of $5\%$ and  $10\%$ of the dataset, respectively. The human accuracy for the classification was $84.70\%$, ranging from $83\%$ to $87\%$.

\subsection{Evaluation}
We keep our evaluation of classification performance consistent with the original paper about the MAMI dataset. \textit{Sub-task A} is evaluated using macro-average F1 measure for each class label (misogynous and not misogynous). Likewise, \textit{Sub-task B} is evaluated using weighted-average F1 measure, weighted by the true label count for each label. For the Hateful Meme dataset, we compare the models based on the macro-average F1-score between the hateful and not-hateful classes. Table \ref{tab:baseline} shows performance statistics for all the participants in the MAMI task within the SemEval-2022 competition~\cite{fersini2022semeval}.  

\begin{table}[!t]
    \centering
    \caption{Classification Head parameters for the Example-based method.}
    \label{tab:classification-head}
        \begin{tabular}{| l | c | c | c| c | c |} 
            \hline
             \textbf{Layers} & \textbf{Dimension} & \textbf{Activation}\\ [0.5ex] 
            \hline
            \textbf{L1} & Feature length * 512  & ReLU\\
            \textbf{L2} & 512 * 256  & ReLU\\
            \textbf{L3} & 256 * 128  & ReLU\\
            \textbf{Prediction} & 128 * Label count  & Sigmoid\\
            \hline 
        \end{tabular}
\end{table}

In addition, we manually evaluate the example-based explanation approach using the visualization tool by analyzing the prediction and similar memes from the training dataset. We evaluate the prototype-based explanation method (xDNN) by its classification performance and manually investigating the prototypes identified from the training dataset.

\begin{table*}[!t]
    \centering
    \caption{MAMI dataset characteristics.}
    \label{tab:dataset}
        \begin{tabular}{| l | c | c | c | c| c | c |} 
            \hline
             \textbf{Sets} & \textbf{Total} & \textbf{Misogynous} & \textbf{Shaming} & \textbf{Stereotype} & \textbf{Objectification} & \textbf{Violence}\\ [0.5ex] 
            \hline
            \textbf{Training} & 10,000 & 5,000 & 1,274 & 2,810 & 2,202 & 953\\
            \hline 
            \textbf{Test} & 1,000 & 500 & 146 & 350 & 348 & 153\\
            \hline 
        \end{tabular}

\end{table*}

\begin{table*}[!t]
    \centering
    \caption{Basic statistics of the results for the participating systems in Sub-task A and Sub-task B, expressed in terms of macro-averaged and weighted-average F1-score respectively.}
    \label{tab:baseline}
        \begin{tabular}{| l | c | c | c | c| c | c | c | c|} 
            \hline
             & \textbf{Evaluation Metric} & \textbf{Min} & \textbf{Q1} & \textbf{Mean} & \textbf{Median} & \textbf{Std Dev} & \textbf{Q3} & \textbf{Max} \\ [0.5ex] 
            \hline
            \textbf{Sub-task A} & macro-averaged F1 & 0.481 & 0.649 & 0.680 & 0.679 & 0.064 & 0.722 & 0.834\\
            \hline 
            \textbf{Sub-task B} & weighted-average F1-score & 0.467 & 0.634 & 0.663 & 0.680 & 0.059 & 0.706 & 0.731\\
            \hline 
        \end{tabular}

\end{table*}

\begin{table*}[!t]
    \centering
    \small
        \caption{Classification results for MAMI dataset and Hateful memes dataset.}
    \label{tab:results-mami-hate}
        \begin{tabular}{| p{10em} | p{10em} | p{10em} | p{10em}| p{7em} |} 
            \hline
            \textbf{Explanation Method} & \textbf{Pretrained Model} & \textbf{Subtask A: Misogyny detection} & \textbf{Subtask B: Misogyny type classification} & \textbf{Hateful Memes} \\ 
            \hline
            \multirow{4}{8em}{\textbf{Prototype-based} (xDNN)} & & & & \\
            & \textbf{BERT$_{Base}$} & 0.537 & 0.524 & 0.485\\
            & \textbf{BERTweet} & 0.543 & 0.534 & 0.445\\ 
            & \textbf{CLIP} & 0.642 & 0.629 & 0.540 \\ 
            & \textbf{CLIP + BERTweet} & 0.648 & 0.626 & 0.541\\ [0.5ex]
            \hline
            \multirow{4}{8em}{\textbf{Example-based} (Neural Classification Head)} & & & & \\
            & \textbf{BERT$_{Base}$} & 0.602 & 0.589 & 0.521 \\
            & \textbf{BERTweet} & 0.600 & 0.594 & 0.503\\ 
            & \textbf{CLIP} & 0.685 & 0.686 & 0.557\\
            & \textbf{CLIP + BERTweet} & \textbf{0.701} & \textbf{0.688} & \textbf{0.583} \\ 
            \hline 
        \end{tabular}
\end{table*}

\subsection{Model Training Details}

The classification model (Figure \ref{fig:classification-model}) used in the example-based explanation setup applies a trainable neural head over frozen pre-trained models, which is trained with the Binary Cross Entropy Loss using the Adam optimizer with a learning rate of $10^{-4}$. Table ~\ref{tab:classification-head} describes each layer of the classification head, and the hidden state of \textbf{L3} is used for feature extraction for similar example searches over the training dataset.

xDNN ~\cite{https://doi.org/10.48550/arxiv.1912.02523} is a generative model, i.e., it learns prototypes and respective distributions automatically from the training data with no user/problem specific parameters. We reuse the publicly available xDNN     implementation\footnote{\url{https://github.com/Plamen-Eduardo/xDNN---Python}} and experiment with different pre-trained models described in the subsection on Pretrained Models. 

\begin{figure*}[!t]
\centering
\includegraphics[width=0.88\linewidth]{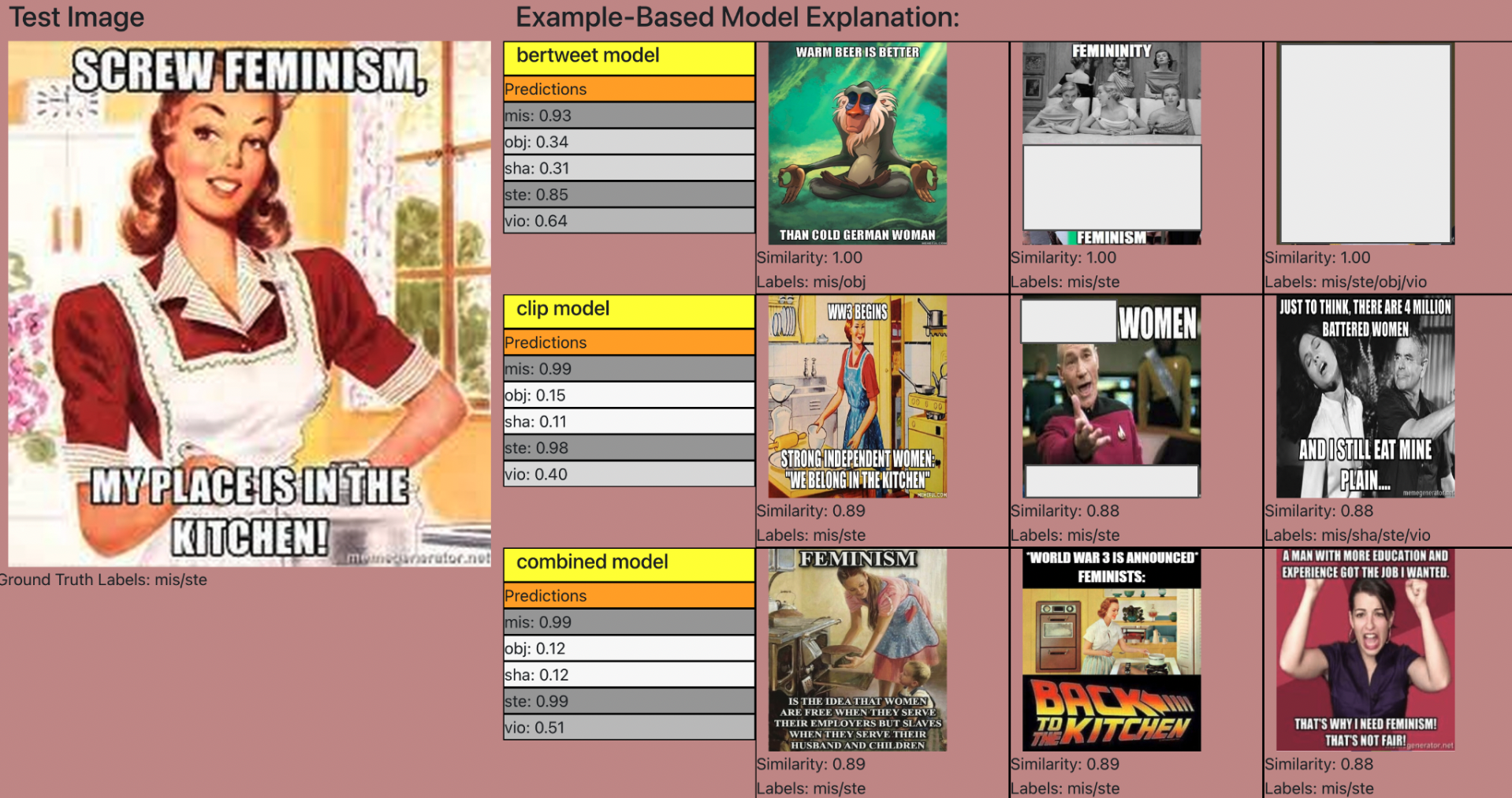}
\caption{Explanatory interface for our Example-based classification method.}
\label{fig:vtool1}
\end{figure*}


\section{Analysis}

In this section, we analyze and compare different methods and feature extraction models for meme classification.
As our methods focus on both accuracy and explainability, we present an analysis on both of them in turn. 


\subsection{Results}

Table ~\ref{tab:results-mami-hate} shows the performance of individual models on the MAMI and Hateful memes datasets. The table compares the explainability strategies' performance over different pre-trained models' choices. 

\textbf{MAMI v/s Hateful Meme:} Between the subtask A (misogyny detection) of the MAMI dataset and Hate detection over the Hateful Meme dataset, all models have better performance for the misogyny detection task. This can be because of the fact that the presence of misogyny directly relates to the mention of women (or related terminology); in contrast, hate is a more open-ended problem. For both tasks and methods, we also observe that the BERTweet model, which is trained on Twitter data, performs better than the BERT-based models for the MAMI dataset, though the difference between the two models is within one point difference. This shows that exposure to social media (Twitter) data has a positive, yet limited, impact on models for meme content classification. However, for the Hateful meme task, BERT performs better than the BERTweet model suggesting the shift in distribution between the two datasets.

\textbf{Prototype-based (xDNN) v/s Example-based Explanation (Deep learning):}  For both datasets and different pre-training models, the Example-based method, which uses a neural classification head, performs better than the  Prototype-based (xDNN) on the same pre-trained model. This indicates that the prototype-based models rely entirely on the pre-trained features and might lose performance on learning complex patterns, which the deep learning model can learn. However, in terms of training time, xDNN is much faster than training the neural classifier head, as it needs just a single pass over the training dataset.

\textbf{Modality performance analysis (Text v/s Image v/s Mixed):} For each meme dataset and method combination, the CLIP-based image model performs comparatively better than BERT-based text models. The combined model using CLIP and BERTweet features outperforms all models, including those using CLIP alone. However, the improvement of the joint model over CLIP is relatively low (0.5-1.5 absolute points) compared to the improvement over BERT models (10-10 absolute points) for both explanation strategies. This means that either visual information is more important than text or that the CLIP model can also capture the textual information in the meme or a combination of both reasons. 
For the task of misogyny classification, the combination of CLIP and BERTweet also performs the best, with the CLIP-only model performing very closely as a second best.

\subsection{Explanability Analysis}

\textbf{Example-Based Classification} Figure~\ref{fig:vtool1} shows the most similar IMs retrieved by the example-based classification for one test image, visualized by our custom-made user interface. The interface displays the model-wise predictions for BERTTweet, CLIP, and their combination, together with similar memes from the training dataset for explainability. The test image is misogynous, portraying a certain stereotype about women. The predictions from each model are correct with high confidence about the misogyny detection and stereotype classification, which is explainable to some degree by looking at similar examples from the training dataset. Focusing on the most similar images per model, we observe that the combined model retrieves three images that also depict misogyny (red background) and stereotyping. The interplay between the text and the vision components is consistent across these IMs, two of which refer to the relationship between the kitchen and women within the context of feminist discourse. This example shows that the instances retrieved by the combined multimodal model are the most reliable, which also correlates with its best performance. 
The examples by CLIP and BERTTweet are partially useful, with CLIP retrieving more relevant images than BERTTweet in most cases. However, in both image-only and text-only settings, the retrieved memes are not exactly related to the test image and are only related to the same overall topic, namely, feminism, or stereotyping. Although CLIP is retrieving images that are related to the test image, even in more fine-grained aspects that demonstrate the textual encoding abilities of CLIP to some extent, it is beneficial to combine the features of CLIP with a dedicated language model that understands slang, like BERTTweet. Focusing on the text-setting only, although the model is performing well, we can observe that none of the memes are accurately about the subject of discussion in the test image. This supports our argument that IM classification should be evaluated both in terms of accuracy and explainability simultaneously, as models might predict correctly for the wrong reasons. 
\textbf{Prototype-based Classification}
xDNN model is inherently explainable as it predicts the label for a meme based on the closest prototype as shown in ~\ref{fig:xDNN}. To our surprise, xDNN for both the MAMI and the Hateful Meme datasets creates prototypes equal to the training supports for the class. This can be because of the fact that the memes, even though belonging to the same category, can have very different textual/visual information content and representation. Nevertheless, we are further investigating this behavior in depth to find the exact reason.



\section{Discussion}

Our experiments revealed that methods for meme classification can balance the goals of explainability and good accuracy. While the example-based method is simpler, it achieved higher performance and its explanatory power was more intuitive, as demonstrated through our tool-supported analysis. Among the feature extractors, we observed that vision models were more effective than language models, and the combination of the two achieved the best performance. We next elevate these analyses to high-level lessons and discussion points that may benefit future work on explainable meme classification for tasks like hate speech detection and misogyny classification.


\textbf{Lack of virtual and real-world context} 
During our analysis, we found that memes often rely on real-world, day-to-day context to be understood. As the pre-trained models lack this context, they may wrongly classify a meme that is very contextual to real-world common sense, especially in social media and misogyny. 
Our models have no access to this additional context. For instance, the usage of the meme can be occurring in an exchange between parties on social media / messaging platforms, which is not provided as a context in the task. A hollistic method for meme classification and explaination needs to account for the cultural and folkloric nature of memes~\cite{Atran2001}: memes often start from a seed of a concept or an idea (this real-world common sense or common cultural background / reference), which then gets derived to achieve an intent. The way this idea is adapted or degenerated depends on the intent and can happen by using figures of speech such as an oxymoron, in which case the image and the meme's text will be antonymous.

\textbf{Integration of background knowledge and figures of speech}
The example in Figure \ref{fig:vtool1} includes a test image where the canvas image depicts a woman in a kitchen. The stereotype and misogyny, in this case, are most likely linked to assumed background knowledge, such as the women's social status in 1960, the second wave of feminism, and the more expansive link between housewives and the kitchen. Although the general topic of discussion can be understood even from the mere reliance on the meme's text, the complete picture and the implicit references are clear, only taking into account both image and text, as is apparent from the examples extracted by models. Particularly, the most similar meme in terms of content and references, the central image in the bottom row has explicit references to World War 3 and the discussion revolving around two opposing opinions: (1) many Gen Z and Millennial women worried about being drafted, and (2) women wanting equality only until they have to be a part of the draft and joining the military. Moreover, by focusing on the other retrieved IMs, you can observe references to various sources. For instance, in the center IM in the last row, you can observe the title of Back to the Future movie that substitutes "future" with "kitchen", implying the relation between these two terms. In the center IM in the middle row, we can observe the Annoyed Picard, the Star Trek character that has long been associated with implying irritability or disappointment that is also extended here towards women. While these cues are captured to some degree by the combined model, still, a need for background commonsense, factual knowledge, as well as internet folklore to build robust and explainable meme classification methods in the future can be seen. This lack of knowledge can be noticed in other extracted images as well. Although they seem to be relevant at first sight and might help in identifying the image as an image containing misogyny, the exact to-the-point reference can still be missing.


\textbf{Subjectivity of Ground Truth Labels} 
Another observation that came out of our analysis is that the problem statement of misogyny detection and type classification is inherently subjective, and the labeler's background and familiarity with social norms affect it. This argument is also clear from the inter-annotator agreement score for the MAMI dataset. Hence, for subjective problem statements like these, the ground truth labels are always questionable and are biased by the labelers. This observation relates to recent work that highlights issues with crowdsourced labeling for hate speech detection or sentiment analysis on social media, as examples~\cite{Morrow2020}, ~\cite{Waseem2017} or~\cite{Davidson2019} to address the bias in labeling hate or abuse datasets. It also connects to discussions of evaluating toxicity~\cite{Carton2020}.
Concerns about the consistency of labeling aspects such as sexism are also prominent in computational social sciences like psychology~\cite{Samory2021}.

\section{Related Work}


Most prior works on Internet memes in AI have focused on understanding their virality and spread on social media over time~\cite{marino2015semiotics,taecharungroj2014effect,ling2021dissecting}. 
Another popular direction has been detecting forms of hate speech in memes.
The Hateful Memes Challenge and Dataset~\cite{kiela2021hateful} is a competition and open source dataset with over 10 thousand examples, where the goal is to leverage vision and language understanding to identify memes that employ hate speech. 
Kirk et al.~\cite{kirk2021memes} compare memes in this challenge to memes in the `wild', observing that extraction of captions is an open challenge, and that open-world memes are more diverse than traditional memes.
The Multimedia Automatic Misogyny Identification (MAMI)~\cite{fersini2022semeval} challenge asks systems to identify misogynous memes, based on both text and images in the input memes. 
Methods for these challenges typically employ Transformer-based models that incorporate vision and language, like ViLBERT~\cite{lu2019vilbert}, UNITER~\cite{chen2020uniter}, and CLIP~\cite{radford2021learning}.
The work by Sheratt~\cite{sherratttowards} aims to organize memes into a genealogy, with the goal of building a comprehensive knowledge base going forward. The combination of efforts to explain IMs with explicit knowledge and the generalization power of large visual, textual, and multimodal models holds a promise to advance the SOTA of meme understanding and classification.
However, to our knowledge, no prior work has focused on such multi-faceted and explainable methods for understanding IMs. To bridge this gap, we design a modular architecture that integrates visual and textual models with prototype- and example-based reasoning methods. Our framework thus balances the goals of obtaining SOTA performance and providing transparent access to the model reasoning. 

There has been a surge in using example-based explanations to enhance people's comprehension of black-box deep learning models' behavior and acquired knowledge. 
\cite{10.1145/3301275.3302289} propose and evaluate two kinds of example-based
explanations in the visual domain. The extracted similar training data points help the end-users understand and recognize the capabilities of the model better.
Although \cite{https://doi.org/10.48550/arxiv.2009.06349} have the same conclusion as \cite{10.1145/3301275.3302289} confirming the effect of examples to boost the comprehension of the model by end-users, they do not see any evidence supporting the same effect about the trust of end-users when presented with example-based explanations. Similarly, methods for prototype-based classification have been developed for visual tasks in the past, such as xDNN~\cite{https://doi.org/10.48550/arxiv.1912.02523}. However, to our knowledge, we are the first work to employ example-based and prototype-based methods for downstream tasks of IM classification.

\section{Conclusions}

In this work, we implemented and analyzed example- and prototype-based approaches for explainable Misogyny Identification and Hate Speech Detection in IMs. 
Our experiments revealed that methods for IMs classification can balance the goals of explainability and good accuracy. While the example-based method is simpler, it achieved higher performance and its explanatory power was more intuitive, as demonstrated through our tool-supported analysis. Among the feature extractors, we observed that vision models were more effective than language models, and the combination of the two achieved the best performance. We connected these findings to thorny challenges about including background knowledge and real-world context in complex multimodal tasks, as well as concerns about the subjective nature of tasks that revolve around these tasks. We make our code available in hope that subsequent research can help us in pursuing these challenges together.


\section*{Acknowledgements}

The first two authors have been supported by armasuisse Science and Technology, Switzerland under contract No. 8003532866. 

\bibliography{meme}

\begin{thebibliography}{53}
\providecommand{\natexlab}[1]{#1}

\bibitem[{Adamic, Lento, and Ng(2014)}]{adamic2014evolution}
Adamic, E. A.~L.; Lento, T.; and Ng, P. 2014.
\newblock The evolution of memes on facebook.
\newblock \emph{Facebook Data Science}.

\bibitem[{Andrews, Diederich, and Tickle(1995)}]{ANDREWS1995373}
Andrews, R.; Diederich, J.; and Tickle, A.~B. 1995.
\newblock Survey and critique of techniques for extracting rules from trained
  artificial neural networks.
\newblock \emph{Knowledge-Based Systems}, 8(6): 373--389.
\newblock Knowledge-based neural networks.

\bibitem[{Angelov and Soares(2019)}]{https://doi.org/10.48550/arxiv.1912.02523}
Angelov, P.; and Soares, E. 2019.
\newblock Towards Explainable Deep Neural Networks (xDNN).

\bibitem[{Atran(2001)}]{Atran2001}
Atran, S. 2001.
\newblock The trouble with memes.
\newblock \emph{Human Nature}, 12: 351--381.

\bibitem[{Badawy, Ferrara, and Lerman(2018)}]{badawy2018analyzing}
Badawy, A.; Ferrara, E.; and Lerman, K. 2018.
\newblock Analyzing the digital traces of political manipulation: The 2016
  Russian interference Twitter campaign.
\newblock In \emph{2018 IEEE/ACM international conference on advances in social
  networks analysis and mining (ASONAM)}, 258--265. IEEE.

\bibitem[{Bessi and Ferrara(2016)}]{bessi2016social}
Bessi, A.; and Ferrara, E. 2016.
\newblock Social bots distort the 2016 US Presidential election online
  discussion.
\newblock \emph{First monday}, 21(11-7).

\bibitem[{Biermann, Kanie, and Kim(2017)}]{biermann2017global}
Biermann, F.; Kanie, N.; and Kim, R.~E. 2017.
\newblock Global governance by goal-setting: the novel approach of the UN
  Sustainable Development Goals.
\newblock \emph{Current Opinion in Environmental Sustainability}, 26: 26--31.

\bibitem[{Cai, Jongejan, and Holbrook(2019)}]{10.1145/3301275.3302289}
Cai, C.~J.; Jongejan, J.; and Holbrook, J. 2019.
\newblock The Effects of Example-Based Explanations in a Machine Learning
  Interface.
\newblock In \emph{Proceedings of the 24th International Conference on
  Intelligent User Interfaces}, IUI '19, 258–262. New York, NY, USA:
  Association for Computing Machinery.
\newblock ISBN 9781450362726.

\bibitem[{Carton, Mei, and Resnick(2020)}]{Carton2020}
Carton, S.; Mei, Q.; and Resnick, P. 2020.
\newblock Feature-Based Explanations Don't Help People Detect
  Misclassifications of Online Toxicity.
\newblock \emph{Proceedings of the International AAAI Conference on Web and
  Social Media}, 14: 95--106.

\bibitem[{Chen et~al.(2022)Chen, Jiang, Chang, Muric, Ferrara
  et~al.}]{chen2022charting}
Chen, E.; Jiang, J.; Chang, H.-C.~H.; Muric, G.; Ferrara, E.; et~al. 2022.
\newblock Charting the information and misinformation landscape to characterize
  misinfodemics on social media: COVID-19 infodemiology study at a planetary
  scale.
\newblock \emph{Jmir Infodemiology}, 2(1): e32378.

\bibitem[{Chen et~al.(2020)Chen, Li, Yu, El~Kholy, Ahmed, Gan, Cheng, and
  Liu}]{chen2020uniter}
Chen, Y.-C.; Li, L.; Yu, L.; El~Kholy, A.; Ahmed, F.; Gan, Z.; Cheng, Y.; and
  Liu, J. 2020.
\newblock Uniter: Universal image-text representation learning.
\newblock In \emph{European conference on computer vision}, 104--120. Springer.

\bibitem[{Davidson, Bhattacharya, and Weber(2019)}]{Davidson2019}
Davidson, T.; Bhattacharya, D.; and Weber, I. 2019.
\newblock Racial Bias in Hate Speech and Abusive Language Detection Datasets.
\newblock 25--35. Association for Computational Linguistics.

\bibitem[{Davison(2012)}]{davison20129}
Davison, P. 2012.
\newblock 9. The Language of Internet Memes.
\newblock In \emph{The social media reader}, 120--134. New York University
  Press.

\bibitem[{Delisle et~al.(2019)Delisle, Kalaitzis, Majewski, de~Berker, Marin,
  and Cornebise}]{delisle2019large}
Delisle, L.; Kalaitzis, A.; Majewski, K.; de~Berker, A.; Marin, M.; and
  Cornebise, J. 2019.
\newblock A large-scale crowdsourced analysis of abuse against women
  journalists and politicians on Twitter.
\newblock \emph{arXiv preprint arXiv:1902.03093}.

\bibitem[{Devlin et~al.(2019)Devlin, Chang, Lee, and
  Toutanova}]{devlin-etal-2019-bert}
Devlin, J.; Chang, M.-W.; Lee, K.; and Toutanova, K. 2019.
\newblock {BERT}: Pre-training of Deep Bidirectional Transformers for Language
  Understanding.
\newblock In \emph{Proceedings of the 2019 Conference of the North {A}merican
  Chapter of the Association for Computational Linguistics: Human Language
  Technologies, Volume 1 (Long and Short Papers)}, 4171--4186. Minneapolis,
  Minnesota: Association for Computational Linguistics.

\bibitem[{DiResta and Grossman(2019)}]{diresta2019potemkin}
DiResta, R.; and Grossman, S. 2019.
\newblock Potemkin pages \& personas: Assessing GRU online operations,
  2014-2019.
\newblock \emph{White Paper https://fsi-live. s3. us-west-1. amazonaws.
  com/s3fs-public/potemkin-pagespersonas-sio-wp. pdf}.

\bibitem[{Fersini et~al.(2022)Fersini, Gasparini, Rizzi, Saibene, Chulvi,
  Rosso, Lees, and Sorensen}]{fersini2022semeval}
Fersini, E.; Gasparini, F.; Rizzi, G.; Saibene, A.; Chulvi, B.; Rosso, P.;
  Lees, A.; and Sorensen, J. 2022.
\newblock SemEval-2022 Task 5: Multimedia automatic misogyny identification.
\newblock In \emph{Proceedings of the 16th International Workshop on Semantic
  Evaluation (SemEval-2022)}, 533--549.

\bibitem[{Ford, Kenny, and
  Keane(2020)}]{https://doi.org/10.48550/arxiv.2009.06349}
Ford, C.; Kenny, E.~M.; and Keane, M.~T. 2020.
\newblock Play MNIST For Me! User Studies on the Effects of Post-Hoc,
  Example-Based Explanations \&; Error Rates on Debugging a Deep Learning,
  Black-Box Classifier.

\bibitem[{Habib and Nithyanand(2022)}]{Habib_Nithyanand_2022}
Habib, H.; and Nithyanand, R. 2022.
\newblock Exploring the Magnitude and Effects of Media Influence on Reddit
  Moderation.
\newblock \emph{Proceedings of the International AAAI Conference on Web and
  Social Media}, 16(1): 275--286.

\bibitem[{Johansen and Kruschke(2005)}]{Johansen2005-nu}
Johansen, M.~K.; and Kruschke, J.~K. 2005.
\newblock Category representation for classification and feature inference.
\newblock \emph{J. Exp. Psychol. Learn. Mem. Cogn.}, 31(6): 1433--1458.

\bibitem[{Khan(2021)}]{khan_2021}
Khan, I. 2021.
\newblock Report of the Special Rapporteur on the Promotion and Protection of
  the Right to Freedom of Opinion and Expression.
\newblock Undocs.org/en/A/76/258. Accessed: 2022-11-30.

\bibitem[{Kiela et~al.(2020)Kiela, Firooz, Mohan, Goswami, Singh, Ringshia, and
  Testuggine}]{https://doi.org/10.48550/arxiv.2005.04790}
Kiela, D.; Firooz, H.; Mohan, A.; Goswami, V.; Singh, A.; Ringshia, P.; and
  Testuggine, D. 2020.
\newblock The Hateful Memes Challenge: Detecting Hate Speech in Multimodal
  Memes.

\bibitem[{Kiela et~al.(2021)Kiela, Firooz, Mohan, Goswami, Singh, Ringshia, and
  Testuggine}]{kiela2021hateful}
Kiela, D.; Firooz, H.; Mohan, A.; Goswami, V.; Singh, A.; Ringshia, P.; and
  Testuggine, D. 2021.
\newblock The Hateful Memes Challenge: Detecting Hate Speech in Multimodal
  Memes.
\newblock arXiv:2005.04790.

\bibitem[{Kirk et~al.(2021)Kirk, Jun, Rauba, Wachtel, Li, Bai, Broestl,
  Doff-Sotta, Shtedritski, and Asano}]{kirk2021memes}
Kirk, H.~R.; Jun, Y.; Rauba, P.; Wachtel, G.; Li, R.; Bai, X.; Broestl, N.;
  Doff-Sotta, M.; Shtedritski, A.; and Asano, Y.~M. 2021.
\newblock Memes in the wild: Assessing the generalizability of the hateful
  memes challenge dataset.
\newblock \emph{arXiv preprint arXiv:2107.04313}.

\bibitem[{Ling et~al.(2021)Ling, AbuHilal, Blackburn, De~Cristofaro, Zannettou,
  and Stringhini}]{ling2021dissecting}
Ling, C.; AbuHilal, I.; Blackburn, J.; De~Cristofaro, E.; Zannettou, S.; and
  Stringhini, G. 2021.
\newblock Dissecting the meme magic: Understanding indicators of virality in
  image memes.
\newblock \emph{Proceedings of the ACM on Human-Computer Interaction},
  5(CSCW1): 1--24.

\bibitem[{Liu et~al.(2019)Liu, Ott, Goyal, Du, Joshi, Chen, Levy, Lewis,
  Zettlemoyer, and Stoyanov}]{https://doi.org/10.48550/arxiv.1907.11692}
Liu, Y.; Ott, M.; Goyal, N.; Du, J.; Joshi, M.; Chen, D.; Levy, O.; Lewis, M.;
  Zettlemoyer, L.; and Stoyanov, V. 2019.
\newblock RoBERTa: A Robustly Optimized BERT Pretraining Approach.

\bibitem[{Lu et~al.(2019)Lu, Batra, Parikh, and Lee}]{lu2019vilbert}
Lu, J.; Batra, D.; Parikh, D.; and Lee, S. 2019.
\newblock Vilbert: Pretraining task-agnostic visiolinguistic representations
  for vision-and-language tasks.
\newblock \emph{Advances in neural information processing systems}, 32.

\bibitem[{Luceri, Cresci, and Giordano(2021)}]{luceri2021social}
Luceri, L.; Cresci, S.; and Giordano, S. 2021.
\newblock Social Media against Society.
\newblock \emph{The Internet and the 2020 Campaign}, 1.

\bibitem[{Luceri et~al.(2019)Luceri, Deb, Badawy, and Ferrara}]{luceri2019red}
Luceri, L.; Deb, A.; Badawy, A.; and Ferrara, E. 2019.
\newblock Red bots do it better: Comparative analysis of social bot partisan
  behavior.
\newblock In \emph{Companion proceedings of the 2019 World Wide Web
  conference}, 1007--1012.

\bibitem[{Luceri, Giordano, and Ferrara(2020)}]{luceri2020detecting}
Luceri, L.; Giordano, S.; and Ferrara, E. 2020.
\newblock Detecting troll behavior via inverse reinforcement learning: A case
  study of russian trolls in the 2016 us election.
\newblock In \emph{Proceedings of the international AAAI conference on web and
  social media}, volume~14, 417--427.

\bibitem[{Mami{\'e}, Horta~Ribeiro, and West(2021)}]{mamie2021anti}
Mami{\'e}, R.; Horta~Ribeiro, M.; and West, R. 2021.
\newblock Are anti-feminist communities gateways to the far right? evidence
  from reddit and youtube.
\newblock In \emph{13th ACM Web Science Conference 2021}, 139--147.

\bibitem[{Marino(2015)}]{marino2015semiotics}
Marino, G. 2015.
\newblock Semiotics of spreadability: A systematic approach to Internet memes
  and virality.

\bibitem[{Medin and Schaffer(1978)}]{Medin1978-je}
Medin, D.~L.; and Schaffer, M.~M. 1978.
\newblock Context theory of classification learning.
\newblock \emph{Psychol. Rev.}, 85(3): 207--238.

\bibitem[{Morrow et~al.(2020)Morrow, Swire-Thompson, Polny, Kopec, and
  Wihbey}]{Morrow2020}
Morrow, G.; Swire-Thompson, B.; Polny, J.; Kopec, M.; and Wihbey, J. 2020.
\newblock The Emerging Science of Content Labeling: Contextualizing Social
  Media Content Moderation.
\newblock \emph{SSRN Electronic Journal}, 40: 3--22.

\bibitem[{Nguyen, Vu, and Tuan~Nguyen(2020)}]{nguyen-etal-2020-bertweet}
Nguyen, D.~Q.; Vu, T.; and Tuan~Nguyen, A. 2020.
\newblock {BERT}weet: A pre-trained language model for {E}nglish Tweets.
\newblock In \emph{Proceedings of the 2020 Conference on Empirical Methods in
  Natural Language Processing: System Demonstrations}, 9--14. Online:
  Association for Computational Linguistics.

\bibitem[{Nogara et~al.(2022)Nogara, Vishnuprasad, Cardoso, Ayoub, Giordano,
  and Luceri}]{nogara2022disinformation}
Nogara, G.; Vishnuprasad, P.~S.; Cardoso, F.; Ayoub, O.; Giordano, S.; and
  Luceri, L. 2022.
\newblock The Disinformation Dozen: An Exploratory Analysis of Covid-19
  Disinformation Proliferation on Twitter.
\newblock In \emph{14th ACM Web Science Conference 2022}, 348--358.

\bibitem[{Pierri, Luceri, and Ferrara(2022)}]{pierri2022does}
Pierri, F.; Luceri, L.; and Ferrara, E. 2022.
\newblock How Does Twitter Account Moderation Work? Dynamics of Account
  Creation and Suspension During Major Geopolitical Events.
\newblock \emph{arXiv preprint arXiv:2209.07614}.

\bibitem[{Pierri et~al.(2022)Pierri, Perry, DeVerna, Yang, Flammini, Menczer,
  and Bryden}]{pierri2022online}
Pierri, F.; Perry, B.~L.; DeVerna, M.~R.; Yang, K.-C.; Flammini, A.; Menczer,
  F.; and Bryden, J. 2022.
\newblock Online misinformation is linked to early COVID-19 vaccination
  hesitancy and refusal.
\newblock \emph{Scientific reports}, 12(1): 1--7.

\bibitem[{Radford et~al.(2021{\natexlab{a}})Radford, Kim, Hallacy, Ramesh, Goh,
  Agarwal, Sastry, Askell, Mishkin, Clark, Krueger, and
  Sutskever}]{https://doi.org/10.48550/arxiv.2103.00020}
Radford, A.; Kim, J.~W.; Hallacy, C.; Ramesh, A.; Goh, G.; Agarwal, S.; Sastry,
  G.; Askell, A.; Mishkin, P.; Clark, J.; Krueger, G.; and Sutskever, I.
  2021{\natexlab{a}}.
\newblock Learning Transferable Visual Models From Natural Language
  Supervision.

\bibitem[{Radford et~al.(2021{\natexlab{b}})Radford, Kim, Hallacy, Ramesh, Goh,
  Agarwal, Sastry, Askell, Mishkin, Clark et~al.}]{radford2021learning}
Radford, A.; Kim, J.~W.; Hallacy, C.; Ramesh, A.; Goh, G.; Agarwal, S.; Sastry,
  G.; Askell, A.; Mishkin, P.; Clark, J.; et~al. 2021{\natexlab{b}}.
\newblock Learning transferable visual models from natural language
  supervision.
\newblock In \emph{International Conference on Machine Learning}, 8748--8763.
  PMLR.

\bibitem[{Renkl(2014)}]{https://doi.org/10.1111/cogs.12086}
Renkl, A. 2014.
\newblock Toward an Instructionally Oriented Theory of Example-Based Learning.
\newblock \emph{Cognitive Science}, 38(1): 1--37.

\bibitem[{Renkl, Hilbert, and Schworm(2009)}]{renkl2009example}
Renkl, A.; Hilbert, T.; and Schworm, S. 2009.
\newblock Example-based learning in heuristic domains: A cognitive load theory
  account.
\newblock \emph{Educational Psychology Review}, 21(1): 67--78.

\bibitem[{Rosch(1973)}]{ROSCH1973328}
Rosch, E.~H. 1973.
\newblock Natural categories.
\newblock \emph{Cognitive Psychology}, 4(3): 328--350.

\bibitem[{Samory et~al.(2021)Samory, Sen, Kohne, Flöck, and
  Wagner}]{Samory2021}
Samory, M.; Sen, I.; Kohne, J.; Flöck, F.; and Wagner, C. 2021.
\newblock “Call me sexist, but...” : Revisiting Sexism Detection Using
  Psychological Scales and Adversarial Samples.
\newblock \emph{Proceedings of the International AAAI Conference on Web and
  Social Media}, 15: 573--584.

\bibitem[{Shao et~al.(2018)Shao, Ciampaglia, Varol, Yang, Flammini, and
  Menczer}]{shao2018spread}
Shao, C.; Ciampaglia, G.~L.; Varol, O.; Yang, K.-C.; Flammini, A.; and Menczer,
  F. 2018.
\newblock The spread of low-credibility content by social bots.
\newblock \emph{Nature communications}, 9(1): 1--9.

\bibitem[{Sherratt(2022)}]{sherratttowards}
Sherratt, V. 2022.
\newblock Towards Contextually Sensitive Analysis of Memes: Meme Genealogy and
  Knowledge Base.

\bibitem[{Sigler(2022)}]{benefits-example-based-explanation}
Sigler, I. 2022.
\newblock Example-based explanations to build better AI/ML Models.

\bibitem[{Starbird, Arif, and Wilson(2019)}]{Starbird2019}
Starbird, K.; Arif, A.; and Wilson, T. 2019.
\newblock Disinformation as collaborative work: Surfacing the participatory
  nature of strategic information operations.
\newblock \emph{Proceedings of the ACM on Human-Computer Interaction}, 3.

\bibitem[{Taecharungroj and Nueangjamnong(2014)}]{taecharungroj2014effect}
Taecharungroj, V.; and Nueangjamnong, P. 2014.
\newblock The effect of humour on virality: The study of Internet memes on
  social media.
\newblock In \emph{7th International Forum on Public Relations and Advertising
  Media Impacts on Culture and Social Communication. Bangkok, August}.

\bibitem[{Tahmasbi et~al.(2021)Tahmasbi, Schild, Ling, Blackburn, Stringhini,
  Zhang, and Zannettou}]{tahmasbi2021go}
Tahmasbi, F.; Schild, L.; Ling, C.; Blackburn, J.; Stringhini, G.; Zhang, Y.;
  and Zannettou, S. 2021.
\newblock “Go eat a bat, Chang!”: On the Emergence of Sinophobic Behavior
  on Web Communities in the Face of COVID-19.
\newblock In \emph{Proceedings of the web conference 2021}, 1122--1133.

\bibitem[{Waseem et~al.(2017)Waseem, Davidson, Warmsley, and
  Weber}]{Waseem2017}
Waseem, Z.; Davidson, T.; Warmsley, D.; and Weber, I. 2017.
\newblock Understanding Abuse: A Typology of Abusive Language Detection
  Subtasks.
\newblock 78--84. Association for Computational Linguistics.

\bibitem[{Zannettou et~al.(2019{\natexlab{a}})Zannettou, Caulfield,
  De~Cristofaro, Sirivianos, Stringhini, and
  Blackburn}]{zannettou2019disinformation}
Zannettou, S.; Caulfield, T.; De~Cristofaro, E.; Sirivianos, M.; Stringhini,
  G.; and Blackburn, J. 2019{\natexlab{a}}.
\newblock Disinformation warfare: Understanding state-sponsored trolls on
  Twitter and their influence on the web.
\newblock In \emph{Companion proceedings of the 2019 world wide web
  conference}, 218--226.

\bibitem[{Zannettou et~al.(2019{\natexlab{b}})Zannettou, Caulfield, Setzer,
  Sirivianos, Stringhini, and Blackburn}]{zannettou2019let}
Zannettou, S.; Caulfield, T.; Setzer, W.; Sirivianos, M.; Stringhini, G.; and
  Blackburn, J. 2019{\natexlab{b}}.
\newblock Who let the trolls out? towards understanding state-sponsored trolls.
\newblock In \emph{Proceedings of the 10th acm conference on web science},
  353--362.

\end{thebibliography}


\end{document}